\documentclass{article}
\usepackage{spconf,amsmath,graphicx}

\usepackage{subcaption}

\DeclareMathOperator*{\argmin}{\operatorname{\arg\min}}
\DeclareMathOperator*{\argmax}{\operatorname{\arg\max}}

\usepackage{amssymb}
\usepackage{amsmath}
\usepackage{dsfont}

\usepackage{comment}

\usepackage{algorithm}
\usepackage{algpseudocode}
\usepackage{array}
\usepackage{alltt}

\usepackage[normalem]{ulem}

\usepackage{makecell}

\usepackage{mathtools, stmaryrd}
\usepackage{xparse} \DeclarePairedDelimiterX{\Iintv}[1]{\llbracket}{\rrbracket}{\iintvargs{#1}}
\NewDocumentCommand{\iintvargs}{>{\SplitArgument{1}{,}}m}
{\iintvargsaux#1} %
\NewDocumentCommand{\iintvargsaux}{mm} {#1\mkern1.5mu,\dots,\mkern1.5mu#2}

\usepackage{amsthm}
\newtheorem{definition}{Definition}[section]
\newtheorem{theorem}{Theorem}[section]

\newtheorem{lemma}[theorem]{Lemma}
\usepackage{xcolor}

\title{The 20 questions game to distinguish Large Language Models}
\name{Gurvan Richardeau, Erwan Le Merrer, Camilla Penzo, Gilles Tredan}
\address{ENS Paris-Saclay, PEReN, INRIA, IRISA, LAAS-CNRS}

\begin{document}
%\ninept
%
%New commands
\newcommand{\indep}{\perp \!\!\! \perp}
\newcommand{\ONE}{\mathds{1}}
\newcommand{\DEF}{\overset{\textbf{def}}{=}}
\newcommand*{\logeq}{\ratio\Leftrightarrow}
\newcommand*{\IID}{\overset{\text{iid}}{\sim}}

% Notations

\newcommand{\Qa}{Q_{\text{audit}}}
\newcommand{\Exp}[1]{\mathcal{#1}_{\text{exp}}}
\newcommand{\Train}[1]{\mathcal{#1}_{\text{Train}}}

\newcommand{\infdiv}{D_\textrm{KL}\infdivx}

\maketitle
\abstract{In a parallel with the 20 questions game, we present a method to determine whether two large language models (LLMs), placed in a black-box context, are the same or not. The goal is to use a small set of (benign) binary questions, typically under 20. We formalize the problem and first establish a baseline using a random selection of questions from known benchmark datasets, achieving an accuracy of nearly 100\% within 20 questions. After showing optimal bounds for this problem, we introduce two effective questioning heuristics able to discriminate 22 LLMs by using half as many questions for the same task. %Nevertheless the generalizability of this improvement has to be studied while random questioning surely generalizes well.
These methods offer significant advantages in terms of stealth and are thus of interest to auditors or copyright owners facing suspicions of model leaks.

\section{Introduction}
\label{sec:intro}

Auditing AI models in a black-box interaction scheme is a difficult task attracting a considerable attention today \cite{tan2018distill,guidotti2018survey}. Although there is now an important set of works for auditing machine learning classifiers (whether for fairness \cite{saleiro2018aequitas}, security \cite{oh2019towards}, intellectual property matters \cite{le2020adversarial} or others), the relative novelty of large language models (LLMs) calls for adapted auditing methods. In this paper, we tackle the possibility of assessing the simple, yet challenging question of efficiently differentiating LLMs only by prompting them. This task has important applications such as being a preamble for a potential prosecution (in the case of model stealing suspicion, see e.g. \cite{defenses} for the same task with classifier models), or evaluating the convergence of these models towards a high accuracy on predefined set of regulatory prompts (thus leading in hardness to discriminate them).% \gilles{If room I'd expand a bit}

In this paper, and after similar attempts in the domain of image classification, \cite{FBI} for example, we take the step to address the issue using \textit{benign} questions, i.e. questions that are not \textit{adversarially} crafted \cite{kumar2023certifying}. These benign inputs have the advantage of being less prone to defenses on the model side \cite{defenses}, as they follow the genuine data distribution.% \erwan{GT comment here:}

\section{Problem Statement}

%\subsection{Definitions and Notations}
We denote $\mathcal{Q}$ the \textit{infinite} set of all questions (i.e. prompts or queries) that have binary answers. In this context, we define a \textit{LLM} or a \textit{model} $m$ as a deterministic map in $ \mathcal{M} \DEF \{0, 1\}^{\mathcal{Q}}$. 

Let then $\mu$ denote the \textit{empirical distribution of the models}. Specifically, a representative sample from $\mu$ is obtained by randomly selecting a model from the collection of all existing models, e.g. those available on Hugging Face.

\subsection{The Problem of Distinguishing LLMs}
\label{sec: Pb Statement}
Consider an auditor having a prompt-only access to two such LLMs $(m,m')$. Her objective is to know whether $m$ and $m'$ are the same model or if they differ. If $m$ and $m'$ are indeed the same model, the two would provide identical responses to any given prompt. However, even if $m$ and $m'$ differ, they may provide identical answers to (at least) some of the prompts. In the following, we will formalize the question of whether $m$ and $m'$ are different and the question of which prompts we should select to efficiently distinguish the two given models.

\begin{definition}[Distinguishing LLMs]
    Given two unknown models $m$, $m'$ from $\mathcal{M}$, find a set of $k$ requests that maximizes the probability of correctly rejecting hypothesis $\mathcal{H}_0 : m = m'$ when $\mathcal{H}_1 : m \neq m'$ is true. In other words, accurately distinguish two distinct LLMs within $k$ prompts.
\end{definition}

This involves analyzing the following map:
\begin{align}\label{eq: Pb_statement}
        S: \,  & \mathbb{N}^* \longrightarrow [0,1] \notag
    \\  & k \mapsto \max_{Q \subset \mathcal{Q}, |Q| = k} \text{acc}(Q)
\end{align}
\noindent The accuracy being:
\begin{equation}\label{eq: accuracy}
    \text{acc}(Q) = \mathbb{P}\left(T_Q = 0, \mathcal{H}_0 \right)
     + \mathbb{P}\left(T_Q = 1, \mathcal{H}_1\right),
\end{equation}
\noindent with $T_Q$ being the test such that the two models are detected as equal if and only if they answer the same on every questions of $Q$, i.e.:

%\begin{equation}
$
    T_Q : (m,m') \mapsto
    \begin{cases}
        0 & \text{if } m(Q) = m'(Q) \quad \text{(accept $\mathcal{H}_0$)} \\
        1 & \text{if } m(Q) \neq m'(Q) \quad \text{(reject $\mathcal{H}_0$)}
    \end{cases}
    $
%\end{equation}

Here, the game becomes devising algorithms that find the best sets of questions which solve (\ref{eq: Pb_statement}) for a small number $k$ of questions.

\subsection{Experimental Assumptions}
To compute the accuracy (\ref{eq: accuracy}), we need the probability that the two models are the same, that is, $\mathbb{P}(\mathcal{H}_0)$. This probability is either the result of our suspicion that the two models are different (e.g., \textit{secret retraining} of a stolen model) or our suspicion that the two are the same (e.g., direct use of a stolen model). We set $\mathbb{P}(\mathcal{H}_0) = \frac{1}{2}$, indicating that we have no stronger belief that the models are identical or different.

We note that given the deterministic behaviours of our models, true positives are trivial in our experiment
%\begin{equation}\label{eq: true positive trivial}
$(\mathbb{P}\left(T_Q = 0 \mid \mathcal{H}_0 \right) = 1)$.
%\end{equation}
Therefore, (\ref{eq: Pb_statement}) can be tackled from the true negatives point of view: %, as it follows:  %from (\ref{eq: true positive trivial}) follows that:
\begin{equation}\label{eq: argmax simplification}
    \argmax_{Q \in \mathcal{Q}, |Q| = k} \text{acc}(Q) 
    = \argmax_{Q \in \mathcal{Q}, |Q| = k} \mathbb{P}\left(T_Q = 1 \mid \mathcal{H}_1 \right). 
\end{equation}
We therefore focus our analysis on the case $m\neq m'$.

\section{Optimal Number of Questions}\label{sec: upper bound}
We define a query as optimal when it separates a set of models into two equal groups according to their answers. If the two groups are evenly divided, the number of differentiated pairs is maximized with a single question (see Appendix \ref{appendix: Proofs} for proof). Let $X$ be the number of queries necessary to differentiate a pair of models $(m,m')$ s.t. $m \neq m’$ and s.t. $m$ and $m'$ are randomly drawn uniformly from the infinite set of models $\mathcal{M}$ where the queries are optimal (see Th. \ref{th: L/2}). Mathematically, let $(m,m') \sim \mathcal{U}(\mathcal{M})^{\otimes 2} \text{ s.t. } m \neq m'$.
Let $X = \inf \{k \in \mathbb{N}^* \mid m(\mathcal{Q}^*(k)) \neq m'(\mathcal{Q}^*(k)) \} $
where $\mathcal{Q}^*$ is $\mathcal{Q}$ ordered such that the queries are optimal. 
The law of $X$ is then
$P(X \leq k) = 1 - (\frac{1}{2})^k,  \forall k \in \mathbb{N}^*$ (see Appendix \ref{appendix: Proofs}).

\section{Experimental Setup}
\label{sec: Experimental Approach}

%\subsection{Approximation}\label{sec:approx}
To tackle (\ref{eq: Pb_statement}), we need a set of questions $\mathcal{Q}$ and the ability to sample from the distribution $\mu$. To achieve this, we select 22 models (listed in Figure \ref{fig: t-sne} in Appendix, set we coin $\mathcal{M}$) and $K$ questions (see Table \ref{table: questions datasets}) from HuggingFace. 
%\cam{This set of models $\mathcal{M}$ and the set of questions constitute our sample of $\mathcal{Q}$}. \cam{This seems to me incorrectly formulated, but I might be wrong} % approximation (approximation because not infinite). \gilles{say $K$ is a biased sample of $Q$?}

As seen in (\ref{eq: argmax simplification}), we are interested in the true negatives, which we approximate by Monte Carlo as follows: %\gilles{monte carlo???}

%\begin{equation}\label{Monte Carlo}
$
\hat{\mathbb{P}} \left(T_Q = 1 \mid\mathcal{H}_1 \right) \propto
%\frac{2}{|\mathcal{M}|(|\mathcal{M}|-1)} 
\sum_{\substack{(m, m') \in \mathcal{M}^2, \\ m \neq m'}} \ONE_{m(Q) \neq m'(Q)}.
$
%\end{equation}
\\Finally, we seek for maximizers over \(\mathcal{Q}\) of (\ref{eq: Pb_statement}). 
An optimal, yet non-tractable, algorithm would require looking at as many sets as the binomial coefficient $\binom{|\mathcal{Q}|}{\text{max\_size}}$, where \text{max\_size} is the maximum size of the question set we consider. We will focus experimentally on the set sizes of $\text{max\_size}=20$, making such an optimal algorithm intractable.
We thus propose different \textit{heuristics} in Sect. \ref{sec: Discriminating Heuristics}.

\paragraph*{The Similarity of LLMs on Their Responses}

\begin{figure}[t!]
\centering
  \centering
  \includegraphics[width=0.8\linewidth]{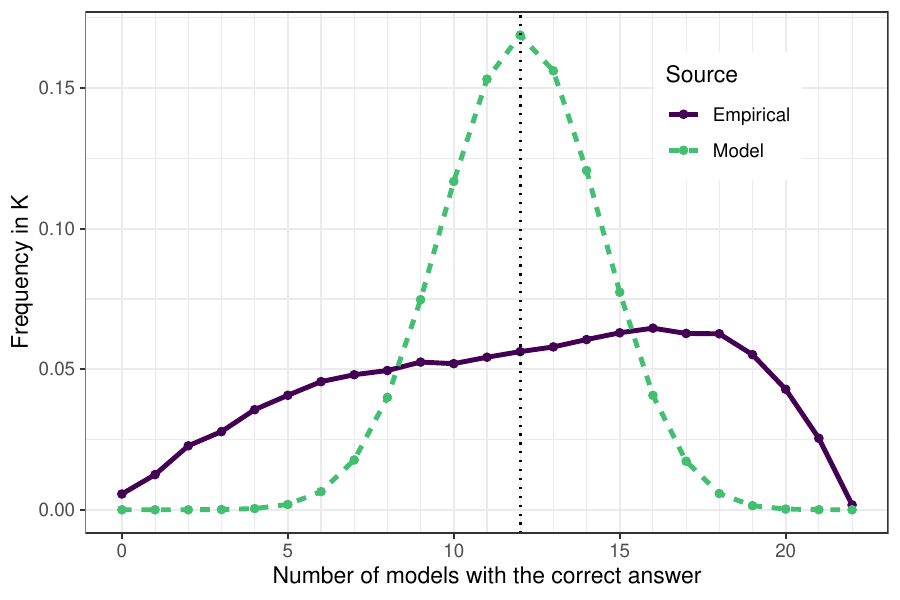}%
  \captionof{figure}{(plain/purple) Distribution of correct answers for questions in $K$, compared to (dashed/green) a random binomial model. Vertical line: average of both distributions. % (5.04).
  }
  \label{fig: ansdist}%
\vspace{-.4cm}
\end{figure}
Figure \ref{fig: ansdist} represents the distribution of models answers for each question of the set $K$. For instance, the point $x=20, y=0.043$ means that 4.3$\%$ of the questions of $K$ were correctly answered by 20 of the 22 tested models. Using the average correct answer rate $(p=54.6\%)$, we present the corresponding binomial distribution (number of success with 22 coin tosses at probability $p$). This model assumes that each LLM answers independently correctly. The resulting bell shaped distribution differs drastically from the empirical observation: % model predicts 0.026$\%$ of the questions are correctly answered by 20/22 models, 163 times less 
this highlights a lack of independence from the models. This means that in practice some questions are easy and correctly answered by most models, while almost all models fail at correctly answering some other (hard) questions.  Motivated by this lack of independence between models, we now look at an experimental approach to identify good question sets.

%\begin{figure}[h!]
%\centering
%\begin{minipage}{.45\linewidth}
%  \centering
%  \includegraphics[width=1.2\textwidth]{figures/%t_sne_on_models_Rule0.png}%
%  \captionof{figure}{t-sne on answer vectors of the models.}%
%  \label{fig: t-sne}%
%\end{minipage}%
%\begin{minipage}{.45\linewidth}
%  \centering%
%    \includegraphics[width=0.9\textwidth]{figures/%model_similarities_on_Dataset_Run_12_Rule0.png}%
%  \captionof{figure}{Model similarities with L1 distance on their %answer vectors. \erwan{A MON AVIS LE TSNE SUFFIT, PAS LA PLACE POUR %LA FIG 2 CORRELATION ON LA COMMENTE POUR LE MOMENT}}%
%  \label{fig:fig: model similarities}%
%\end{minipage}
%\end{figure}

\paragraph*{Experimental method}
We consider heuristic algorithms of the form 
$
\mathcal{A}: (\mathcal{M}, \mathcal{Q}),  \mapsto \lambda \in [0,1]^{\mathcal{Q}}
$
that yields a \textit{discriminating score} $\lambda$ such that for $q \in \mathcal{Q}$, the closer $\lambda(q)$ is to 1, the better $q$ is supposed to distinguish models of $\mu$.

Then, for each pair $(m,m')$ of $\mathcal{M}$ models, we query both models with the questions from $\mathcal{Q}$ ordered decreasingly by $\lambda$, until they answer differently (i.e. we distinguished them) and record the corresponding number of questions. From this we plot the cumulative distribution function that indicates how many model pairs can be distinguished as a function of the number of questions. See Alg. \ref{alg: Experimental Algorithm}, where:
$
t(Q, m, m') = \min_{j \in \Iintv{1, |Q|}} \{ j \mid m(Q[j]) \neq m'(Q[j])\}.
$

\paragraph*{A Baseline: Selecting Questions at Random}
We define the Random heuristic $\mathcal{A}_{\text{rand}}$ as 
%\begin{equation}
$\mathcal{A}_{\text{rand}}(M, Q) = q \mapsto \mathcal{U}([0,1]).$
%\end{equation}
Fig \ref{fig: all CDF in line} (a) illustrates that by randomly selecting questions uniformly across all datasets, we achieve an average accuracy of 95\% with 6 questions. %Also, 20 questions appear here as an empirical upper bound \gilles{upper bound de quoi ? }. 
The best case reached within the 2000 runs (green curve) is based on AUC (Area Under Curve).%\gilles{vraiment pas sur de ce best case}
%Figure \ref{fig: rand multi} presents the same data, but separated by individual datasets. This comparison highlights the discriminating power of each dataset, although the differences are not substantial. In terms of the AUC (Area Under Curve) metric, MedMCQA performs the best.
The figure representing each question dataset is omitted, as the differences are not substantial: achieving 95\% accuracy requires 5 questions for the ``best" dataset vs 7 for the ``worst".

\begin{figure*}[htb]
  \centering
    \begin{subfigure}{0.25\linewidth}
        \includegraphics[width=1\linewidth]{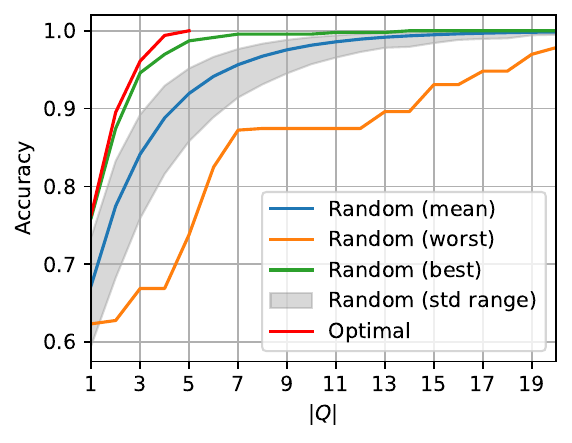}
    \vspace{-.7cm}
        \caption{$\mathcal{A}_\text{rand}$}
        \label{fig:result1}
    \end{subfigure}\hfill
    \begin{subfigure}{0.25\linewidth}
    \includegraphics[width=1.0\linewidth]{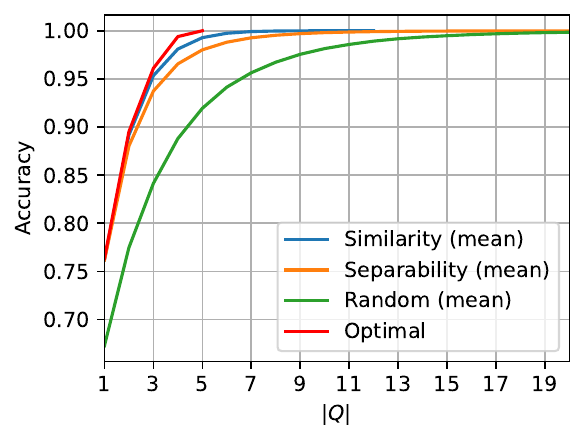}
    \vspace{-.7cm}
    \caption{$\mathcal{A}_\text{rand}$ vs $\mathcal{A}_\text{sep}$ vs $\mathcal{A}_\text{sim}$ (mean)}
    \label{fig:result3}
    \end{subfigure}\hfill
    \begin{subfigure}{0.25\linewidth}
    \includegraphics[width=1.0\linewidth]{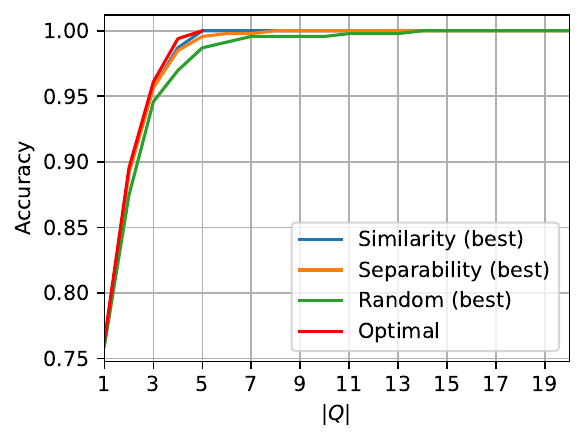}
    \vspace{-.7cm}
    \caption{$\mathcal{A}_\text{rand}$ vs $\mathcal{A}_\text{sep}$ vs $\mathcal{A}_\text{sim}$ (best)}
    \label{fig:result4}
    \end{subfigure}\hfill
  \caption{% lot of the (approximation of the)  %E:je ne comprends pas ce debut de phrase
  Map 
  $S: k \mapsto \max_{Q \in \mathcal{Q}, |Q| = k} \text{acc}(Q)$, for all questions datasets combined and the 22 LLMs. Each heuristic has been run 2000 times, and we present the mean, std, best and worst cases (based on AUC).}\label{fig: all CDF in line}
  \vspace{-.4cm}
\end{figure*}

\section{Two Heuristics for Distinguishability}\label{sec: Discriminating Heuristics}

In the following, $Q \subset \mathcal{Q}$ and $M \subset \mathcal{M}$ denote finite subsets of questions and models. $N$ is a min-max normalization.
%\gilles{I believe we can remove alg1.}

\begin{algorithm}
\scriptsize
    \caption{$\mathcal{A}_\text{sep}$: The Separability Heuristic}
    \label{alg: Experimental Algorithm}
    \begin{algorithmic}[1]
    \Require $\mathcal{M}$, $\mathcal{Q}$, $\mathcal{A}$ (a heuristic).
        \State $\lambda \gets \mathcal{A}(\mathcal{M}, \mathcal{Q})$
        \State $\mathcal{Q}_{\mathcal{A}} \gets \text{sort}(\mathcal{Q}; \text{by decreasing } \lambda)$  
        \State Let $\mathcal{D}$ an empty dict with default values equal to 0.
        \For{ each $(m, m') \in \mathcal{M}^2$, s.t. $m \neq m'$}
            \State $ 
            \text{n\_queries} \gets t(\mathcal{Q}_{\mathcal{A}}, m, m')$
            \State $\mathcal{D}[\text{n\_queries}] \gets \mathcal{D}[\text{n\_queries}] + 1$
        \EndFor
    \State \Return $\mathcal{D}$
    \end{algorithmic}
    \label{alg:separability}
\end{algorithm}
\subsection{The Separability Heuristic}
We here focus on individual questions: what makes it a good question?
An initial measure of distinguishing power is how evenly the binary question divides the two groups of models. The heuristic is given in Algorithm \ref{alg:separability}, here is a formal definition:
\begin{definition}[Subset Separability]\label{def: subset separability}
    %We define the 
    \textbf{Separability} 
    of $ X \subset M $:
{\small
 \begin{align*}
    \Delta_{M}:  & \mathcal{P}(M) \longrightarrow [0,1] \\
   & X \mapsto N \left( |M| - \left| |X| - |\Bar{X}| \right| \right),
\end{align*} %G: ici tu dis que tu suppose card(H) divisible mais tu présente le cas non divisible
    where $\Bar{X} = M \setminus X$.
}
\end{definition}
We denote $M_q \DEF \{m \in M \mid m(q) = 1 \}$ such that the highest $\Delta_M(M_q)$ is, the better $q$ divides the models evenly.
Finally, we define the heuristic $\mathcal{A}_{\text{sep}}$ as follows: 
{\small{
\begin{equation}
\mathcal{A}_{\text{sep}}(M, Q) = q \mapsto 
\begin{cases}
    \mathcal{U}([0,1]) & \text{if } q \in \argmax_{q \in \mathcal{Q}} \Delta_M(M_q) \\
    0 & \text{else}
\end{cases}
\end{equation}
}}
\subsection{The Recursive Similarity Heuristic}
We now consider sets of questions: what makes it a good question set?
Sampling from questions with maximum separability makes it possible for two questions to obtain the same partitions (i.e. the same two groups of models), making it pointless to use them both. Hence, the need for a metric that compares the similarity between the two partitions. We propose a heuristic that constructs a sequence of questions using a recursive approach, ensuring that each subsequent question is as dissimilar as possible from the previous.

\begin{definition}[Similarity of two partitions of same separability]
For $ X, Y \subset M$ s.t. $\Delta_M(X) = \Delta_M(Y)$,
we define the \textbf{similarity} $\Gamma_{M}$ , as:
\small
\begin{align*}
    \Gamma_{M}: & \; \mathcal{P}(M) \times \mathcal{P}(M)\longrightarrow \left[0,1   \right]
    \\ & (X, Y) \mapsto N\left( \max \{ |X \cap Y|, |\Bar{X} \cap Y|, |X \cap{\Bar{Y}}|, |\Bar{X} \cap \Bar{Y}| \} \right)
\end{align*}
%See Appendix \ref{appendix: Similarity Examples} for examples.
\end{definition}

Then we define the heuristic $\mathcal{A}_\text{sim}$ with Algorithm \ref{alg: A_sim} such that
$\mathcal{A}_{\text{sim}}(M, Q) = \text{RecursiveSim}(M,Q, \text{empty list}).$
This algorithm initially selects two questions with the least similarity (line 3), a process that is tractable for two questions. Subsequently, it iteratively adds a question that is the least similar to the existing ones (line 8). Each search for a new question incurs a linear cost, ensuring the algorithm remains tractable. Finally, as this process yields very good discriminating questions, we can define a limit of iterations (\text{max\_iter}) making the algorithm even more efficient (the questions that have not been added have a score set to $0$). 

\begin{algorithm}
\scriptsize
\caption{$\mathcal{A}_\text{sim}$: The Recursive Similarity Heuristic}
    \label{alg: A_sim}
    \begin{algorithmic}[1]
    \Require $M$, $Q$, $Q_\text{sim}$
    \If {$Q_\text{sim}$ is empty}
    \State $Q_\text{sep}^* \gets \argmax_{q \in Q} \Delta_M(M_q)$ 
    \State Let $q_1, q_2 \in \argmin_{(q,q') \in Q^2}\Gamma_M(M_q, M_{q'})$
    \State $Q_\text{sim}$.append$(q_1, q_2)$
    \State \Return RecursiveSim$(M,Q_\text{sep}^*, Q_\text{sim}$) 
    \EndIf
    \If{$len(Q_\text{sim}) \leq \text{max\_iter}$} 
    \State Let $q \in \argmin_{q \in Q \setminus Q_\text{sim}}\left\{
     \sum_{q' \in Q_\text{sim}} \Gamma_M(M_q, M_{q'}) 
     \right\}$ 
    \State $Q_\text{sim}$.append$(q)$
    \State \Return RecursiveSim$(M, Q, Q_\text{sim})$
    \Else 
    \State $\lambda \gets q \mapsto$
    $\begin{cases}
        1 - \frac{\text{rank}(q)}{\text{max\_iter}} & \text{if } q \in Q_\text{sim} \\
        0 & \text{else}
    \end{cases}$
    \State \Comment{$\text{where rank}(q)$ is such that $Q_\text{sim}[\text{rank}(q)] = q$}
    \State \Return $\lambda$
    \EndIf
    
    \end{algorithmic}
\label{alg:recursive}
\end{algorithm}

\section{Experimental Results}
Figure \ref{fig: all CDF in line} (a) shows how the random baseline performs.
Figure \ref{fig: all CDF in line} (b) shows that selecting questions with the highest separability at random produces significantly better results than a purely random selection. It also shows that the $\mathcal{A}_\text{sim}$ heuristic performs better, although with modest improvement, suggesting that our set of questions with maximum separability (in our experiments, we got 4,000 such questions) is quite diverse in the partitions the questions yield.
While achieving $95\%$ accuracy requires on average 6 questions using random selection, our heuristics only require 3. Fig \ref{fig: all CDF in line} (c) represents the best set of questions obtained by each heuristic against our specific set of models.

\begin{figure}[t!]
\centering
  \centering
  \includegraphics[width=0.75\linewidth]{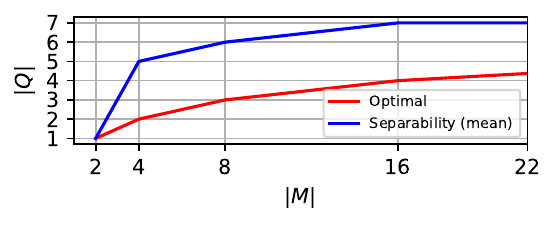}%
\vspace{-.4cm}
  \captionof{figure}{Scalability: number of questions $|Q|$ needed to distinguish 99\% of the pairs of a set of $|M|$ models, for the Separability heuristic and the optimal case.%\erwan{TODO rajouter ou remplacer separability par heuristique recursive}
  }
  \label{fig: t-sne}%
\vspace{-.4cm}
\end{figure}

\vspace{-.4cm}
\section{Conclusion}

Our framework shows that properly selecting from binary questions shows promising results on the task of distinguishing LLMs. This encourages further investigation; in particular a generalization to wider sets of models. %maybe including a train/test split approach. 
Also, focusing on differentiating models that are by construction close to each other would constitute an interesting angle, e.g. models with different training parameters or training data. Furthermore,  investiguating  the robustness of our approach with non-deterministic models is futurework.

\appendix
\vspace{-.4cm}

\section{Datasets and Models}\label{appendix: datasets and models}
%Please refer to Table \ref{table: questions datasets}.
\begin{table}[h!]
\centering
\scriptsize{
\begin{tabular}{c | c | c | c}
\hline
\textbf{\makecell{Dataset}} & \textbf{\makecell{Number of \\ choice}} & \textbf{\makecell{Experiment \\ Subset Size}} & \textbf{\makecell{Original \\ Size}} \\
\hline
hellaswag & 4 & 20k & 40k \\
\hline
MedMCQA & 4 & 20k & 182k \\
\hline
mmlu & 4 & 20k & 115k \\
\hline
piqa & 2 & 16k & 16k \\
\hline
\end{tabular}
}
\caption{Question Datasets for Experiments (details in  \cite{zellers2019hellaswagmachinereallyfinish} \cite{bisk2019piqareasoningphysicalcommonsense}
\cite{hendrycks2021measuringmassivemultitasklanguage}
\cite{pal2022medmcqalargescalemultisubject}).
Note that their cropping was not performed randomly, which may result in a biased representation of the datasets. We binarized the datasets that were not, by mapping to 0 a wrong answer and 1 a correct one.}
\label{table: questions datasets}
\end{table}

\begin{figure}[t!]
\centering
  \centering
  \includegraphics[width=0.8\linewidth]{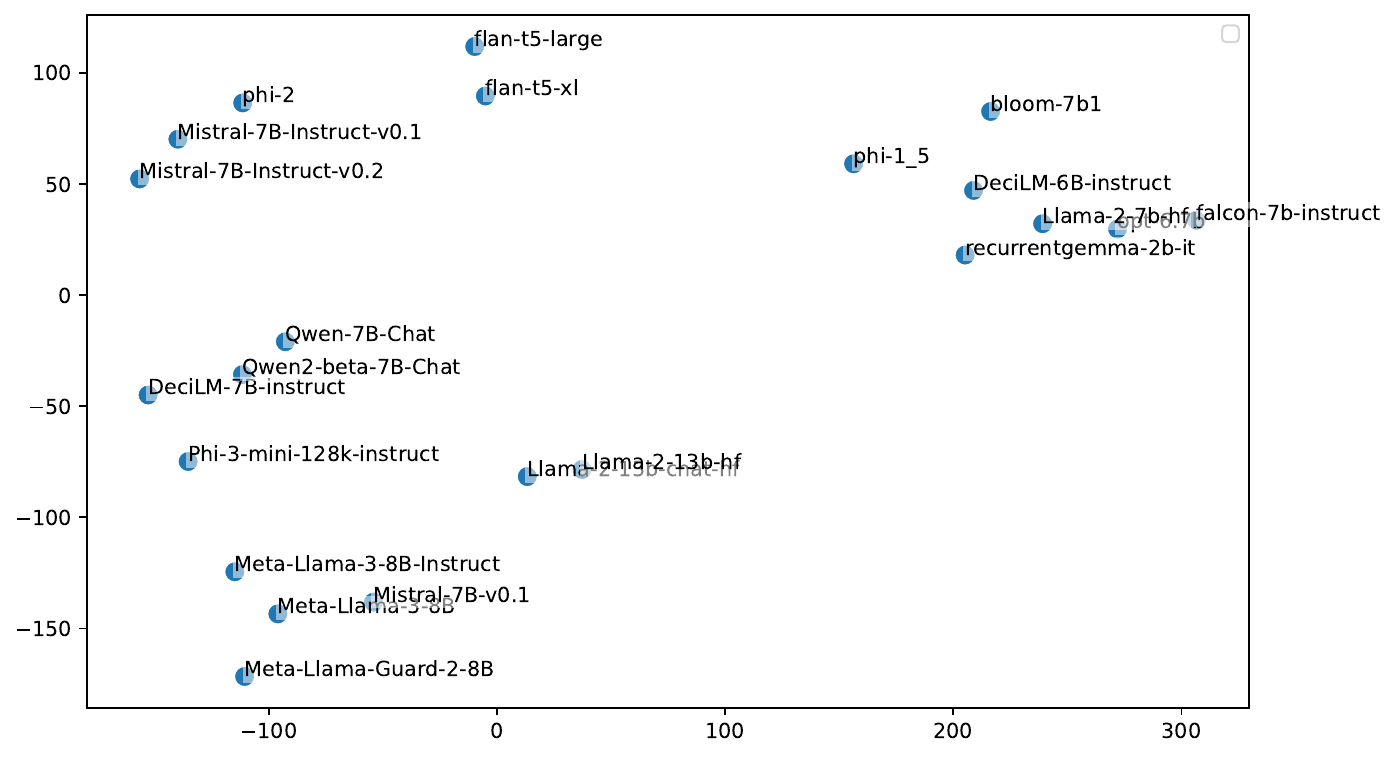}%
  \captionof{figure}{Proximity (t-sne on response vectors) of the 22 LLMs.
  Those of same \textit{family} i.e. from same company and differing by version appear globally close, except the DeciLM models.%DeciLM-6B-instruct and DeciLM-7B-Instruct.
  }
  \label{fig: t-sne}%
\vspace{-.4cm}
\end{figure}

\section{Proofs}\label{appendix: Proofs}
\begin{theorem}\label{th: L/2}
    Let $M$ be a finite set such that $|M| = L$. 
    A question that maximally differentiates pairs in \( M \) is one that splits the set into the most equal groups.
\end{theorem}
\begin{proof}
Let $q_k$ be the query that splits $M$ into two groups of size $ 2 \leq k \leq L - 1$ and $L - k$. We have:
\begin{align*}
|\{\text{pairs split by } q_k\}| &= |\{\text{all pairs}\}| - (|P_1| + |P_2|)  \\
 &=\binom{L}{2} - \left(\binom{k}{2} + \binom{L-k}{2}\right)
\end{align*}
where $P_1, P_2$ are the pairs in the first and second groups.
Next, we aim to maximize the number of split pairs, which are the pairs that are differentiated. Therefore, we seek for the maximizer of $k \mapsto |\{\text{pairs split by } q_k\}|$ that is a second order polynomial with maximum reached in $\left\lfloor \frac{L}{2} \right\rfloor$.
%\begin{align*}
    %&\argmax_{2 \leq k \leq L - 1} \left\{ \binom{L}{2} - \left( \binom{k}{2} + \binom{L-k}{2} \right) \right\} \\
    %=& \argmin_{2 \leq k \leq L - 1} \left\{ k(k-1) + (L-k)(L-k-1) \right\} \\ 
    %=& \argmin_{2 \leq k \leq L - 1} \left( 2k^2 - 2Nk + L(L-1) \right) = \left\lfloor \frac{L}{2} \right\rfloor.
%\end{align*}
\end{proof}

% E: cannot remove a B2 uses it too
%\gilles{I removed reference to this lemma in the optimality section. I think we can remove this lemma }
\begin{lemma}\label{lemma: upperbound_finite}
For a finite set of models $M$ s.t. $|M|=2^n$. The law of $X$ is given by:
\begin{equation}
    \mathbb{P}(X \leq k) = 1 - \frac{2^{n-k} - 1 }{
    2^n -1
    }, \; 1 \leq k \leq n.
\end{equation}
\end{lemma}

\begin{proof}
At first step, i.e. $k=1$, there are $2^n$ models, the first query $\mathcal{Q}^*(1)$ will divide in two groups of size $2^{n-1}$. In each of these groups, there are $\binom{2^{n-1}}{2}$ possible pairs, whereas there are $\binom{2^n}{2}$ different pairs before querying. Therefore, we deduce that $\mathbb{P}(X \leq 1) = \frac{|\{\text{pairs that the query splits} \}|}{|\{\text{all possible pairs of $M$} \}|}
= \frac{\binom{2^n}{2} - 2.\binom{2^n-1}{2}}
{\binom{2^n}{2}}$.
At step $k \in \mathbb{N}^*$, there are $2^{k-1}$ groups of $2^{n-(k-1)}$ models, query $\mathcal{Q}^*(k)$ will divide each of these groups by two. In each of these $2^k$ groups, there are $\binom{2^{n-k}}{2}$ possible pairs.
With similar thinking as first step, we deduce that:
$\mathbb{P}(X \leq k) = \frac{|\{\text{pairs that the queries split} \}|}{|\{\text{all possible pairs of $M$} \}|}
= \frac{\binom{2^n}{2} - 2^k.\binom{2^{n-k}}{2}}
{\binom{2^n}{2}} = 1 - 2^k.\frac{2^{n-k}(2^{n-k} -1)}{2^n(2^n -1)} = 1 - \frac{2^{n-k} - 1 }{2^n -1}.$
\end{proof}

\begin{theorem}\label{Th: law of X infinite}
    For a infinite and countable set of models $M$. The law of $X$ is given by:
    \begin{equation}
        P(X \leq k) = 1 - \left(\frac{1}{2} \right)^k,  \forall k \in \mathbb{N}^*.
    \end{equation}
\end{theorem}

\begin{proof}
    The result is obtained by taking the limit of $n$ towards infinity of $\mathbb{P}(X \leq k)$ of Lemma \ref{lemma: upperbound_finite}. 
\end{proof}

\newpage
\bibliographystyle{IEEEbib}
\bibliography{strings,refs}

\begin{thebibliography}{10}

\bibitem{tan2018distill}
Sarah Tan, Rich Caruana, Giles Hooker, and Yin Lou,
\newblock ``Distill-and-compare: Auditing black-box models using transparent
  model distillation,''
\newblock in {\em Proceedings of the 2018 AAAI/ACM Conference on AI, Ethics,
  and Society}, 2018, pp. 303--310.

\bibitem{guidotti2018survey}
Riccardo Guidotti, Anna Monreale, Salvatore Ruggieri, Franco Turini, Fosca
  Giannotti, and Dino Pedreschi,
\newblock ``A survey of methods for explaining black box models,''
\newblock {\em ACM computing surveys (CSUR)}, vol. 51, no. 5, pp. 1--42, 2018.

\bibitem{saleiro2018aequitas}
Pedro Saleiro, Benedict Kuester, Loren Hinkson, Jesse London, Abby Stevens, Ari
  Anisfeld, Kit~T Rodolfa, and Rayid Ghani,
\newblock ``Aequitas: A bias and fairness audit toolkit,''
\newblock {\em arXiv preprint arXiv:1811.05577}, 2018.

\bibitem{oh2019towards}
Seong~Joon Oh, Bernt Schiele, and Mario Fritz,
\newblock ``Towards reverse-engineering black-box neural networks,''
\newblock {\em Explainable AI: interpreting, explaining and visualizing deep
  learning}, pp. 121--144, 2019.

\bibitem{le2020adversarial}
Erwan Le~Merrer, Patrick Perez, and Gilles Tr{\'e}dan,
\newblock ``Adversarial frontier stitching for remote neural network
  watermarking,''
\newblock {\em Neural Computing and Applications}, vol. 32, no. 13, pp.
  9233--9244, 2020.

\bibitem{defenses}
Daryna Oliynyk, Rudolf Mayer, and Andreas Rauber,
\newblock ``I know what you trained last summer: A survey on stealing machine
  learning models and defences,''
\newblock {\em ACM Comput. Surv.}, vol. 55, no. 14s, jul 2023.

\bibitem{FBI}
Thibault Maho, Teddy Furon, and Erwan Le~Merrer,
\newblock ``Model fingerprinting with benign inputs,''
\newblock in {\em ICASSP 2023 - 2023 IEEE International Conference on
  Acoustics, Speech and Signal Processing (ICASSP)}, 2023, pp. 1--5.

\bibitem{kumar2023certifying}
Aounon Kumar, Chirag Agarwal, Suraj Srinivas, Soheil Feizi, and Hima Lakkaraju,
\newblock ``Certifying llm safety against adversarial prompting,''
\newblock {\em arXiv preprint arXiv:2309.02705}, 2023.

\bibitem{zellers2019hellaswagmachinereallyfinish}
Rowan Zellers, Ari Holtzman, Yonatan Bisk, Ali Farhadi, and Yejin Choi,
\newblock ``Hellaswag: Can a machine really finish your sentence?,'' 2019.

\bibitem{bisk2019piqareasoningphysicalcommonsense}
Yonatan Bisk, Rowan Zellers, Ronan~Le Bras, Jianfeng Gao, and Yejin Choi,
\newblock ``Piqa: Reasoning about physical commonsense in natural language,''
  2019.

\bibitem{hendrycks2021measuringmassivemultitasklanguage}
Dan Hendrycks, Collin Burns, Steven Basart, Andy Zou, Mantas Mazeika, Dawn
  Song, and Jacob Steinhardt,
\newblock ``Measuring massive multitask language understanding,'' 2021.

\bibitem{pal2022medmcqalargescalemultisubject}
Ankit Pal, Logesh~Kumar Umapathi, and Malaikannan Sankarasubbu,
\newblock ``Medmcqa : A large-scale multi-subject multi-choice dataset for
  medical domain question answering,'' 2022.

\end{thebibliography}

\end{document}